\documentclass[]{fairmeta}
\usepackage[utf8]{inputenc}
\usepackage{wrapfig}
\usepackage{tabularx}
\usepackage{textcomp}
\usepackage{stfloats}
\usepackage{url}
\usepackage{verbatim}
\usepackage{titlesec}
\usepackage{tocloft}
\usepackage{adjustbox}
\usepackage{multirow}
\usepackage{pifont}
\usepackage{tikz}
\usepackage{comment}
\usepackage{amsmath,amssymb,amsfonts}
\usepackage{colortbl}
\usepackage{color}
\usepackage{booktabs}
\usepackage{graphicx}
\usepackage{subcaption}
\usepackage{array}
\usepackage{makecell}
\usepackage{overpic}
\usepackage{threeparttable}
\usepackage{float}
\usepackage{xspace}
\usepackage{listings}

\preto\section{\FloatBarrier}

\definecolor{headerbg}{RGB}{240,244,248}
\definecolor{groupbg}{RGB}{240,244,248}
\definecolor{oursbg}{RGB}{234,247,236}
\definecolor{softgreen}{RGB}{234,247,236}
\definecolor{bestbg}{RGB}{235,243,252}
\definecolor{softblue}{RGB}{235,243,252}
\definecolor{lossbg}{RGB}{252,235,235}
\definecolor{softred}{RGB}{252,235,235}
\definecolor{goodgreen}{RGB}{34,139,34}
\definecolor{badred}{RGB}{192,57,43}
\definecolor{softgray}{RGB}{100,100,100}
\definecolor{posgreen}{RGB}{76,175,80}
\definecolor{negyellowgreen}{RGB}{190,215,80}

\renewcommand{\paragraph}[1]{\vspace{1.25mm}\noindent\textbf{#1}}

\setlabdisplayname{OmniAI Group of ZJU ACES Lab}
\setuniversityname{Zhejiang University}

\usepackage{hyperref}
\usepackage{url}
\usepackage{enumitem}
\usepackage{cleveref}

\title{VLA-Rolling: Fewer Denoising Steps, Better Control via Adaptive Action Reuse}

\author[1]{Sicheng He} 
\author[1]{Yi Pan} 
\author[1]{Qiuting Yu}
\author[1]{Jiaming Huang}
\author[1]{Qi Lu}
\author[1]{Shuyan Wu}
\author[1]{Hongyan Feng}
\author[1]{Xuanyu Liu}
\author[1]{Jun Zhang}
\author[1]{Jie Zhang}
\author[1]{Wenqi Zhang}
\author[1]{Xuhong Zhang}
\author[1]{Jianwei Yin}

\affiliation[1]{Zhejiang University}
\metadata[Email]{Sicheng He \email{chen123ww@outlook.com}; Jie Zhang \email{carlzhang4@zju.edu.cn}; Wenqi Zhang \email{zhangwenqi@zju.edu.cn}; Xuhong Zhang \email{zhangxuhong@zju.edu.cn}}
\abstract{
Flow-matching vision-language-action (VLA) models generate an action chunk by iteratively denoising Gaussian noise, but typically execute only a short prefix before replanning with updated observations. Under this receding-horizon paradigm, the unexecuted remainder---the \textbf{action tail}---is discarded, and the next chunk is regenerated from fresh noise. We show that this discarded tail retains \textbf{transferable denoising progress}: action-tail initialization can match or outperform a native 10-step Gaussian baseline using fewer denoising steps. However, the optimal denoising budget varies substantially across samples, indicating that tail reuse should be adapted to the updated environment. Based on this observation, we introduce \textbf{VLA-Rolling}, a lightweight inference framework that reuses action tails through three stages: \textbf{action-tail inheritance}, \textbf{per-action confidence estimation}, and \textbf{confidence-guided refinement} through renoising and denoising budget allocation, while the underlying VLA remains frozen. Across $\pi_{0.5}$, SmolVLA, X-VLA, and MolmoAct2 on LIBERO, MetaWorld, LIBERO-Plus, and RoboTwin 2.0, VLA-Rolling reduces action-chunk inference cost in all 16 model--benchmark pairs, achieving up to a $1.48\times$ speedup, while improving success rate in 14 pairs by up to $15.08$ percentage points. These results show that action tails are not merely discarded predictions, but reusable intermediate states that enable more efficient and accurate VLA inference. These results demonstrate that the action tail can support the generation of better actions with fewer denoising steps. Our project is available at \url{https://chenbao123ww.github.io/VLA-Rolling/}
}

\date{October 2026}

\begin{document}
\thispagestyle{firstheader}
\maketitle
\pagestyle{empty}

\section{Introduction}
\label{sec:introduction}

Flow-matching vision-language-action (VLA) models generate future robot actions from visual observations and language instructions~\citep{black2024pi_0, intelligence2025pi_, shukor2025smolvla, zheng2026x, lee2025molmoact, bjorck2025gr00t, cai2026xiaomi}. Their architectures differ, but they share a flow-matching action generator that iteratively transforms Gaussian noise into a sequence of future actions, called \textbf{action chunk}~\citep{jing2025mixture, liang2026adaptive,xu2026knowing, wang2026vla, lu2026faster}. In models such as $\pi_{0.5}$~\citep{intelligence2025pi_} and SmolVLA~\citep{shukor2025smolvla}, this generator is implemented as a dedicated Transformer module called the \textbf{action expert}, conditioned on features from a vision-language model (VLM). Thanks to this , the robot can plan and execute several coherent steps without calling the model after every action. Yet the whole chunk is predicted from one observation, so environmental changes can make later actions inaccurate. To limit this error accumulation, \textbf{receding-horizon execution} runs only a short prefix, then observes and replans~\citep{zhao2023learning,jing2025mixture,liang2026adaptive}. The price of this feedback is a fresh start: the unexecuted remainder, or \textbf{action tail}~\citep{huang2026chainvla}, is discarded, and the next chunk starts again from Gaussian noise.

A fresh start may waste completed computation. The action tail has already undergone denoising and overlaps in time with the actions needed in the next inference round, so it may be a better starting point than fresh Gaussian noise. Yet the tail was generated from an earlier observation and may have become stale. It is neither a ready-to-execute plan nor fresh noise: its value as a starting point depends on how much the environment has changed. This tension raises a compelling question:

{\centering
\emph{Can we exploit the action tail to generate new actions
\textbf{faster} without sacrificing action \textbf{accuracy}?}\par}

We probe this possibility on $\pi_{0.5}$ and SmolVLA (\Cref{sec:motivating_observations}). On LIBERO and MetaWorld, tail initialization matches or improves the action error of a 10-step Gaussian baseline in four denoising steps for $\pi_{0.5}$ and six for SmolVLA (\Cref{fig:tail_few_step}). Yet no fixed short budget suits every sample: the sample-wise optimal step counts span the full range from 1 to 10 for both models (\Cref{fig:optimal_step_distribution}). The question is therefore not simply whether to reuse the tail, but how much to trust it under each new observation.

\begin{figure}[H]
  \centering
  \includegraphics[width=\linewidth]{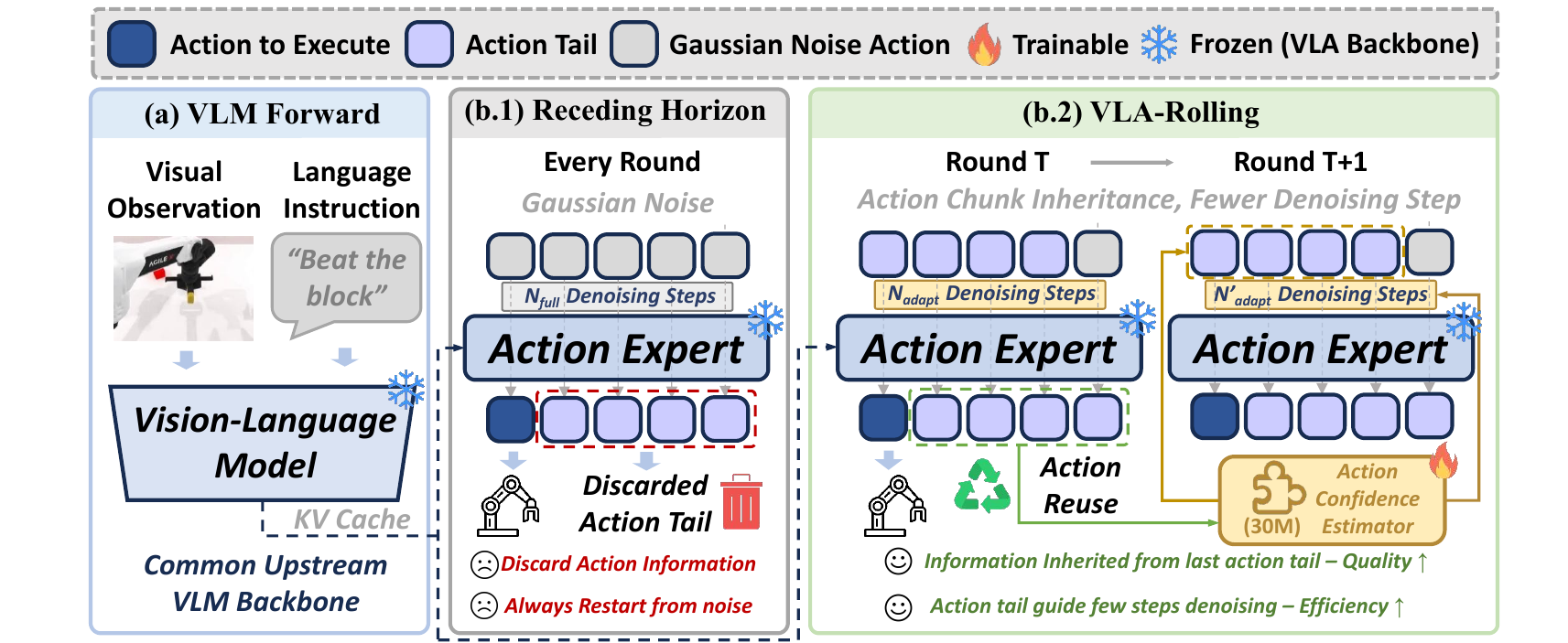}
  \caption{\textbf{VLA-Rolling reuses rather than discards action tails.} Confidence controls both how strongly inherited actions are renoised and how many denoising steps refine the next chunk.}
  \label{fig:intro}
\end{figure}

To adapt reuse to each new observation, we introduce \textbf{VLA-Rolling}, a three-stage inference framework (\Cref{fig:intro}). First, \textbf{action-tail inheritance} aligns unexecuted actions with the next planning window, using completed denoising work as the starting point for a new prediction. Second, a lightweight \textbf{action confidence estimator} predicts the reuse value of each inherited action from the new observation and state change. We train it offline on temporally adjacent demonstrations: smaller error between an inherited action and its aligned demonstrated action yields a higher confidence target, while the loss penalizes overestimated confidence more heavily. Third, \textbf{confidence-guided refinement} injects more noise into low-confidence actions and assigns more denoising steps when the chunk contains an unreliable action. The underlying VLA remains frozen.

To summarize, our main contributions are:
\begin{itemize}[itemsep=1.5pt, topsep=1.5pt, leftmargin=20pt]
    \item[(1)] \textit{Action-Tail Reuse Potential} (\Cref{sec:motivating_observations}): We reveal that discarded tails retain transferable denoising progress: four steps for $\pi_{0.5}$ and six for SmolVLA match a 10-step Gaussian baseline. Yet the sample-wise optimal budgets span 1--10 steps, exposing both the opportunity for reuse and the need to adapt it to each new observation.

    \item[(2)] \textit{Confidence-Guided Rolling Reuse} (\Cref{sec:method}): We propose VLA-Rolling, a lightweight inference framework that learns per-action reuse confidence from temporally adjacent demonstrations. This confidence jointly controls how much of each inherited action to retain and how many denoising steps to allocate, preserving useful motion while correcting stale predictions without retraining the base VLA.

    \item[(3)] \textit{Extensive Experiments and Analysis} (\Cref{sec:experiments}): On $\pi_{0.5}$, SmolVLA, X-VLA, and MolmoAct2 across LIBERO, MetaWorld, LIBERO-Plus, and RoboTwin 2.0, VLA-Rolling reduces action-chunk inference time in all 16 pairs, achieving up to a \textbf{$1.48\times$} speedup (32.52\% lower cost), and improves success in 14 pairs by up to 15.08 percentage points. Ablations expose the risk of naive reuse, while real-world target changes trigger additional refinement.
\end{itemize}

\section{Probing the Feasibility of Action Tail Reuse}
\label{sec:motivating_observations}

Flow-matching VLA jointly predicts an $H$-step action chunk from a single environment state $\mathbf{s}_t$. However, during the execution of the action chunk, the latest observations from the environment are not used to update the action, causing all subsequent actions to be based on the state observed before execution began. Errors in early actions cause a deviation between the actual state of the robot and the environment and the state expected by the remaining predicted actions, and this state mismatch propagates to subsequent actions, gradually accumulating throughout the execution scope.

Receding-horizon execution mitigates such error accumulation by periodically closing the perception--action loop. Instead of executing all $H$ predicted actions, the robot only the first $K$ actions (where $K<H$), then obtains the updated observations and generates a new action block. This execution paradigm consumes more inference iterations in exchange for the ability to use new observations to update actions more frequently.

This execution strategy also leaves part of every generated action chunk unused. After executing the first $K$ actions, the remaining $L=H-K$ actions are
\begin{equation}
    \mathbf{A}^{\mathrm{tail}}_t
    =
    \mathbf{A}_t^{K:H},
    \label{eq:action_tail}
\end{equation}

which we refer to as the \textbf{action tail}. Under conventional receding-horizon execution, this tail is discarded once the observation is refreshed. However, our statistics show that under the execution horizon of common VLA models, the environment does not change significantly. Observations and proprioceptions have high similarity in nearly 98\% of adjacent environment steps. Detail statistical data is shown in the Appendix.~\ref{sec:env_sim}.

\begin{figure}[!htbp]
  \centering
  \includegraphics[
    width=\linewidth
  ]{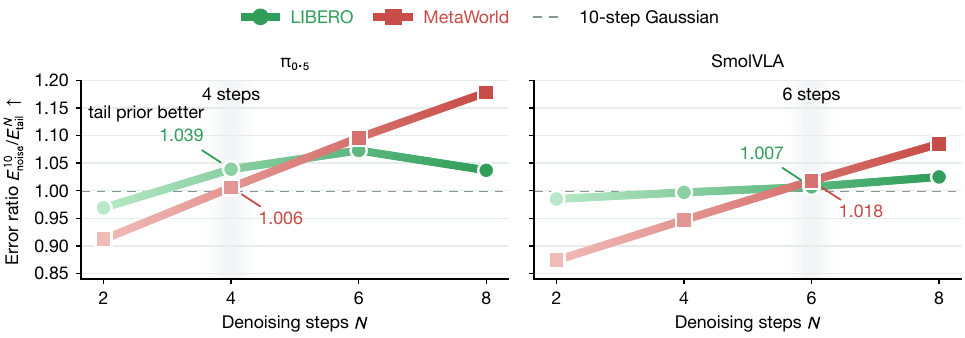}
  \caption{
    \textbf{Action tails recover a 10-step Gaussian prediction in fewer steps.}
    Tail initialization crosses the Gaussian baseline in four steps for $\pi_{0.5}$ and six for SmolVLA on both benchmarks. Subtle shaded bands mark these earliest shared operating points.
  }
  \label{fig:tail_few_step}
\end{figure}

\paragraph{Diagnostic Evaluation Setup.} We conducted diagnostic experiments on $\pi_{0.5}$ and SmolVLA to analyze the properties of the action tail after updating observations. For two time steps $t-K$ and $t$ separated by $K$ control steps in the same demonstration trajectory, we first generate an action chunk of length $H$ using the frozen VLA in the old state $\mathbf{s}_{t-K}$. After assuming the first $K$ actions have been executed, the remaining $L=H-K$ actions constitute the action tail. This tail is temporally aligned with the demonstration action starting at time $t$. Subsequently, in the updated state $\mathbf{s}_t$, we initialize with this action tail and fresh Gaussian noise respectively, and compare them using the same target action $\mathbf{A}_{t}^{*}$.

\paragraph{Diagnostic Evaluation Result.} We first investigate whether it's possible to generate the target action with fewer denoising steps by starting from the sampled action tail. Specifically, we perform $N\in\{2,4,6,8\}$ steps of denoising on each inherited action tail, and compare the final error $E_{\mathrm{tail}}^{N}$ with the error of the complete Gaussian noise denoising result $E_{\mathrm{noise}}^{N_{full}}$.

\Cref{fig:tail_few_step} shows that an inherited tail can surpass the 10-step Gaussian baseline in fewer denoising steps: four for $\pi_{0.5}$ and six for SmolVLA on both benchmarks. However, more steps do not always improve the prediction. For $\pi_{0.5}$ on LIBERO, the error ratio falls from 1.073 at six steps to 1.037 at eight, indicating a larger action error after further refinement. This raises a sample-level question: \emph{How many denoising steps does each inherited tail need?}

We define the sample-wise optimal denoising step as

\begin{equation}
    N^* = \arg\min_{N} E_{\mathrm{tail}}^{N}
\end{equation}

where $N \in \{1,\ldots,N_{full}\}$. We then analyze the distribution of $N^*$ for both models.

\begin{figure}[!htbp]
  \centering
  \includegraphics[
    width=\linewidth
  ]{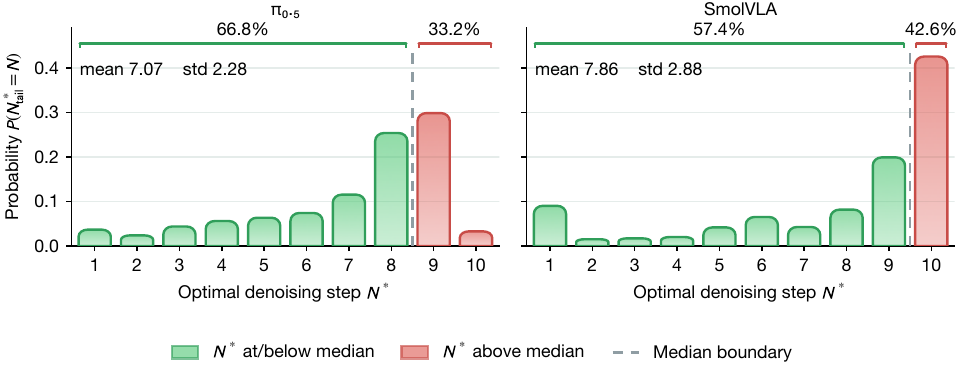}
  \caption{
    \textbf{Action tails require sample-dependent refinement.}
    Each bar shows the fraction of samples with a given optimal denoising step $N^*$. Green denotes $N^*$ at or below the model-specific median (8 for $\pi_{0.5}$, 9 for SmolVLA), while red denotes $N^*$ above it. Dashed lines mark the median boundaries; brackets report the fraction of samples in each region. Means and standard deviations appear inside the panels.
  }
  \label{fig:optimal_step_distribution}
\end{figure}

As shown in \Cref{fig:optimal_step_distribution}, $N^*$ spans every step from 1 to 10, with standard deviations of 2.28 for $\pi_{0.5}$ and 2.88 for SmolVLA. Even beyond the respective medians of 8 and 9 steps, 33.2\% and 42.6\% of samples require further refinement. This variation motivates selecting the denoising budget separately for each sample.

Together, these diagnostics motivate a sample-adaptive budget: reuse the tail when fewer denoising steps suffice, while allowing further refinement when needed.

\section{VLA-Rolling: Adaptive Action Tail Reuse}
\label{sec:method}

\begin{figure}[!htbp]
  \centering
  \includegraphics[
    width=\linewidth
  ]{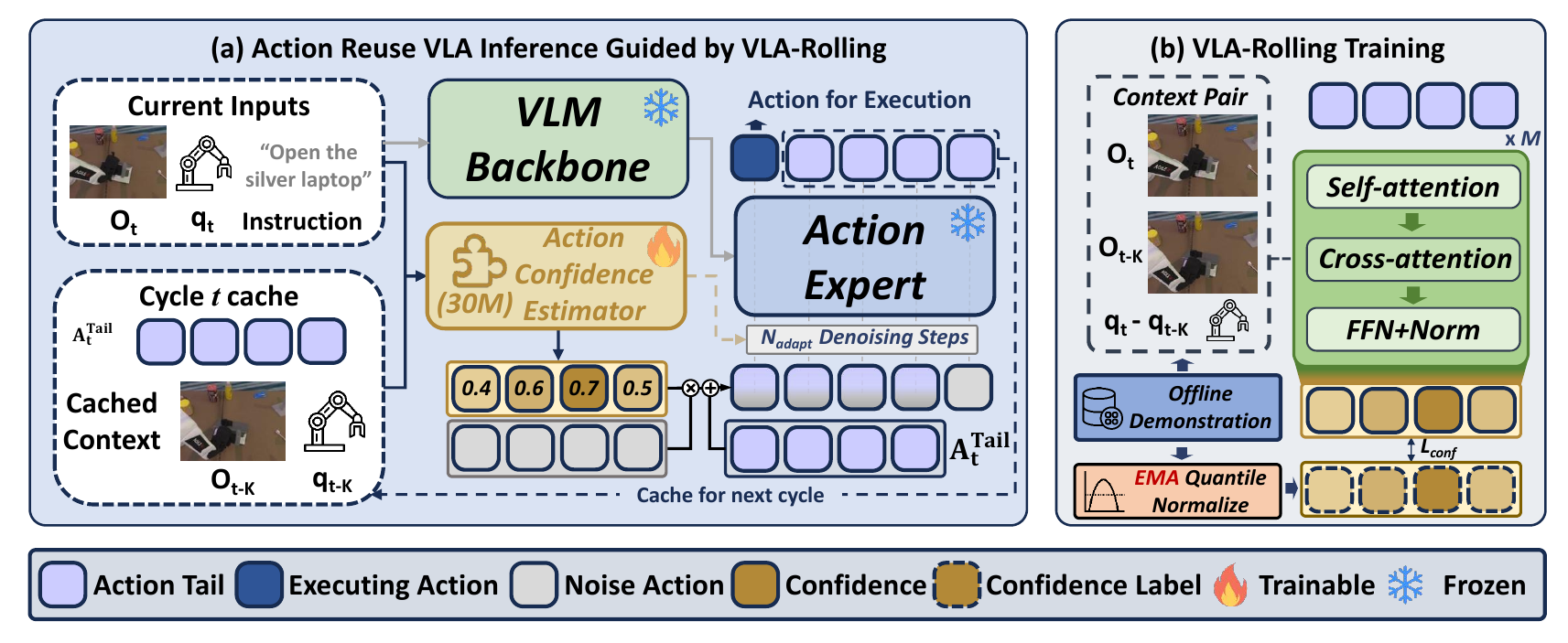}
    \caption{
    \textbf{Overview of VLA-Rolling.}
    VLA-Rolling reuses the unexecuted action tail from the previous inference round by estimating its confidence, which jointly determines tail preservation and the adaptive denoising budget while keeping the original VLA frozen.
    \textbf{(a) VLA-Rolling inference.}
    The Action Confidence Estimator predicts a confidence which controls its interpolation with Gaussian noise. The resulting initialization is refined using $N_{\mathrm{adapt}}$ denoising steps. The executable prefix is sent to the robot, while the remaining tail rolls into the next round.
    \textbf{(b) VLA-Rolling training.}
     Confidence labels are constructed from the quality of inherited actions and normalized with EMA-based quantile statistics.
    }
  \label{fig:VLA-Rolling_overview}
\end{figure}

\textbf{VLA-Rolling} attempts to utilize widely existing but underutilized information: the \textbf{action tail}. VLA-Rolling is a inference paradigm that uses the action tail of each round as prior knowledge for the next inference round, as shown in \Cref{fig:VLA-Rolling_overview}. VLA-Rolling are designed into three phases: \textbf{(1) Action-tail inheritance stage} (\Cref{sec:action_inherit}) is used to retain the action and environment state obtained by the previous model after denoising, which serves as the basis for building the starting point of the current denoising stage; \textbf{(2) Action tail estimation stage} (\Cref{sec:action_confidence}) uses the trained action confidence estimator to evaluate the retention value of the action based on the environment state and its changes; \textbf{(3) Confidence-guided refinement stage} (\Cref{sec:confidence_sampling}) generates the starting point of the action by adding noise and assigns it the corresponding denoising cost based on the obtained action tail and confidence.

\subsection{Action Tail Inheritance}
\label{sec:action_inherit}

During the action inheritance phase, We obtain the preserved action tail from previos round, temporally aligne it with action at time $t$.
\begin{equation}
    \mathbf{A}_{t-K}
    =
    \pi_{\theta}(\mathbf{o}_{t-K},\mathbf{q}_{t-K},\ell),
    \qquad
    \mathbf{A}_{t-K}^{tail}
    =
    \mathbf{A}_{t-K}^{K:H}
    \in\mathbb{R}^{L\times d_a},
    \qquad
    L=H-K,
    \label{eq:diagnostic_tail_construction}
\end{equation}

Meanwhile, for the following phases, we need to use the environment state saved in the previous round, and after obtaining the environment state in this round, calculate the corresponding environmental change:

\begin{equation}
    \Delta \mathbf{o_{t}} = \mathbf{o_{t}} - \mathbf{o_{t-k}}
    \qquad
    \Delta \mathbf{q_{t}} = \mathbf{q_{t}} - \mathbf{q_{t-k}}
\end{equation}

where observation $\mathbf{o}_{t}$ and proprioceptive state $\mathbf{q}_{t}$ denote the current environmental state, $\Delta\mathbf{o}_{t}$ and $\Delta\mathbf{q}_{t}$ denote the environmental change.

\subsection{Action Confidence Estimation}
\label{sec:action_confidence}

To estimate the number of denoising steps required for the current action, we propose the \textbf{Action Confidence Estimator}, a lightweight plugin that predicts the confidence of the action tail based on the current environmental changes and state. We formalize action confidence estimator as $f_{\phi}$:

\begin{equation}
    c_{\phi}
    =
    f_{\phi}
    \left(
        \mathbf{A}_{t-K}^{\mathrm{tail}}
        \,\middle|\,
        \mathbf{o}_t,
        \Delta\mathbf{o}_t,
        \mathbf{q}_t,
        \Delta\mathbf{q}_t
    \right),
    \label{eq:ace_confidence}
\end{equation}

The output $c \in \mathbb{R}^{L\times 1}, $ represents the estimated reliability of retaining the inherited action at corresponding position.

To supervise action confidence estimator to learn representing the reusability of actions under changing environments, we choose the error between the action tail and the target action $E_{t}$ as the representation of behavior in the new environment. This is because the closer the action tail was to the target in the previous denoising process, the easier it is to refine, and therefore it should receive a higher confidence. We then perform a logarithmic transformation on $E_{t}$ to obtain the risk $r_{t}$. Because the action tail is already quite close to the target action, this transformation makes subtle differences in the action more discernible.

\begin{equation}
    E_{t}
    =
    \left\|
        \mathbf{A}_{t-K}^{tail}
        -
        \mathbf{A}^{*}_{t}
    \right\|_{2}^{2},
    \qquad
    r_{t}
    =
    \log(E_{t}),
    \label{eq:ace_risk}
\end{equation}

Given the wide distribution of action errors, we use log transformation to compress larger error ranges while preserving discriminative power in smaller error regions. $r_{t}$ measures the absolute deviation of the inherited action relative to the aligned demonstration action.

Meanwhile, considering that action confidence estimator needs to be adapted to different models and environments, we perform quantile normalization on risk to unify the error scale:

\begin{equation}
    c^{*}_{t}
    =
    1-
    \operatorname{clip}
    \left(
        \frac{
            r_{t}-q_{0.1}
        }{
            q_{0.9}-q_{0.1}
        },
        0,
        1
    \right).
    \label{eq:ace_conf_target}
\end{equation}

Action confidence estimator is trained by predicting the difference between confidence and risk confidence target. Our training objective is formulated as:

\begin{equation}
    \mathcal{L}_{\mathrm{conf}}
    =
    \mathbb{E}_{\mathbf{A}_{t}^{tail},o_{t-K},o_{t},q_{t-K},q_{t}}
    \left[
    \left\|
        w_i\left(c_{\phi}-c^{*}\right)
    \right\|_1
    \right],
    \qquad
    w_i=
    \begin{cases}
        w_{+}, & c_{\phi,i}>c^{*}_{t,i},\\
        w_{-}, & c_{\phi,i}\leq c^{*}_{t,i},
    \end{cases}
    \label{eq:confidence_prediction_loss}
\end{equation}

where $w_{+}>w_{-}$ assigns a larger penalty to overestimating the confidence of an action tail.

\subsection{Confidence-Guided Refinement}
\label{sec:confidence_sampling}

% First, \textbf{action-tail inheritance} aligns unexecuted actions with the next planning window, using completed denoising work as the starting point for a new prediction. Second, a lightweight \textbf{action confidence estimator} predicts the reuse value of each inherited action from the new observation and state change. We train it offline on temporally adjacent demonstrations: smaller error between an inherited action and its aligned demonstrated action yields a higher confidence target, while the loss penalizes overestimated confidence more heavily. Third, \textbf{confidence-guided refinement} injects more noise into low-confidence actions and assigns more denoising steps when the chunk contains an unreliable action. The underlying VLA remains frozen. 

To optimize the inference process through preservable prior information in the action tail, we propose the \textbf{Rolling Action Reuse} inference paradigm. The trained action confidence estimator performs a forward operation before each subsequent action generation to obtain the confidence of the tail of the current inherited action. This confidence then guides two aspects of action generation: $(1)$ Renoising determines how much Gaussian noise should be injected as annealing before action generation to reduce the likelihood of actions going out of distribution; $(2)$ Adaptive-Step Denoising determines the denoising budget allocated for action generation.

The prefix of the action start point is determined by renoising. We will use the predicted confidence as the noise linear interpolation coefficient for the action tail at different positions:

\begin{equation}
    \mathbf{A}^{\mathrm{init}}_{t,i}
    =
    \begin{cases}
        c_{\phi,i}\mathbf{A}_{t-K,i}^{tail}+(1-c_{\phi,i})\boldsymbol{\epsilon},
        & 0\leq i<L,\\
        \boldsymbol{\epsilon}_i,
        & L\leq i<H.
    \end{cases}
    \label{eq:confidence_renoising}
\end{equation}

Despite confidence-dependent annealing of the inherited action tail, the positions without an inherited counterpart are supplied with Gaussian initialization. High-confidence positions retain more temporal information and receive less noise. Low-confidence positions receive stronger renoising, approaching the native VLA initialization for greater regeneration under the current context when the inherited action is unreliable.

During the adaptive-step denoising stage, the confidence level serves as the basis for determining the number of denoising steps for the entire action chunk:
\begin{equation}
    c_{\min}=\min_{0\leq i<L}c_{\phi,i},
    \qquad
    N_{\mathrm{adapt}}
    =
    \operatorname{clip}
    \left(\left\lceil(1-c_{\min})N_{full}\right\rceil, N_{\min}, N_{full}\right),
    \label{eq:adaptive_budget}
\end{equation}

where $N_{\min}\geq1$ and $N_{full}$ bound the budget, equaling to the default denoising steps of the frozen VLA.

In this paradigm, the action tail of each inference round is not discarded, but rather rolled forward temporally. By the time the action is finally to execute, it has already been continuously refined over multiple rounds of rolling.

\section{Experiments}
\label{sec:experiments}

We tested on LIBERO~\citep{liu2023libero}, MetaWorld~\citep{yu2020meta}, LIBERO-Plus~\citep{fei2025libero} and RoboTwin 2.0~\citep{chen2025robotwin}, using $\pi_{0.5}$ as the main VLA backbone, and SmolVLA, X-VLA, and MolmoAct2 as cross-comparison targets. Our experiments were organized into four parts: \textbf{(1) Overall performance} comparisons showing the model's baseline performance on various benchmarks and its performance after deploying VLA-Rolling; and \textbf{(2) Ablation experiments} verifying the impact of different components in VLA-Rolling on inference results.;\textbf{(3) Analysis Experiments}; \textbf{(4) Real-world experiments} verifying the deployment of VLA-Rolling on real-world VLA models. 

\subsection{Overall Performance}

\begin{table*}[t]
\centering
\caption{
Performance and inference efficiency comparison with VLA-Rolling.
\textbf{SR} denotes the average success rate across tasks, and
\textbf{Cost} denotes the inference time per action chunk. \textbf{Bold} data indicates an improvement compared to the baseline.
}
\label{tab:VLA-Rolling_results}

\small
\setlength{\tabcolsep}{2.2pt}   % 原来 3.5pt，减小列间距
\renewcommand{\arraystretch}{1.0}

\begin{tabular}{@{}l*{8}{c}@{}}
\toprule
\multirow{2}{*}{Model}
& \multicolumn{2}{c}{\textbf{LIBERO}}
& \multicolumn{2}{c}{\textbf{MetaWorld}}
& \multicolumn{2}{c}{\textbf{LIBERO-Plus}}
& \multicolumn{2}{c}{\textbf{RoboTwin}} \\
\cmidrule(lr){2-3}
\cmidrule(lr){4-5}
\cmidrule(lr){6-7}
\cmidrule(lr){8-9}
& SR (\%)$\uparrow$ & Cost (ms)$\downarrow$
& SR (\%)$\uparrow$ & Cost (ms)$\downarrow$
& SR (\%)$\uparrow$ & Cost (ms)$\downarrow$
& SR (\%)$\uparrow$ & Cost (ms)$\downarrow$ \\
\midrule

$\pi_{0.5}$
& 96.38 & 168.09
& 67.22 & 222.69
& 84.73 & 167.65
& 66.54 & 172.92 \\
\textbf{+VLA-Rolling}
& \textbf{97.83} & \textbf{129.60}
& \textbf{67.88} & \textbf{184.21}
& \textbf{85.42} & \textbf{147.40}
& \textbf{67.26} & \textbf{164.27} \\
$\Delta$ (pp/\%)
& \cellcolor{posgreen!18}\textbf{+1.45}
& \cellcolor{posgreen!62}\textbf{-22.90\%}
& \cellcolor{posgreen!55}\textbf{+0.66}
& \cellcolor{posgreen!49}\textbf{-17.28\%}
& \cellcolor{posgreen!14}\textbf{+0.69}
& \cellcolor{posgreen!37}\textbf{-12.08\%}
& \cellcolor{posgreen!14}\textbf{+0.72}
& \cellcolor{posgreen!20}\textbf{-5.00\%} \\
\midrule

X-VLA
& 96.03 & 135.24
& 57.08 & 123.31
& \textbf{66.04} & 136.80
& \textbf{60.68} & 131.09 \\
\textbf{+VLA-Rolling}
& \textbf{96.65} & \textbf{112.79}
& \textbf{61.46} & \textbf{90.05}
& 65.91 & \textbf{121.93}
& 60.40 & \textbf{96.99} \\
$\Delta$ (pp/\%)
& \cellcolor{posgreen!14}\textbf{+0.62}
& \cellcolor{posgreen!48}\textbf{-16.60\%}
& \cellcolor{posgreen!35}\textbf{+4.38}
& \cellcolor{posgreen!70}\textbf{-26.97\%}
& \cellcolor{negyellowgreen!25}-0.13
& \cellcolor{posgreen!34}\textbf{-10.87\%}
& \cellcolor{negyellowgreen!30}-0.28
& \cellcolor{posgreen!68}\textbf{-26.01\%} \\
\midrule

SmolVLA
& 73.85 & 127.80
& 65.84 & 122.88
& 20.74 & 130.90
& 25.36 & 148.68 \\
\textbf{+VLA-Rolling}
& \textbf{79.78} & \textbf{89.57}
& \textbf{70.74} & \textbf{102.57}
& \textbf{35.82} & \textbf{120.97}
& \textbf{31.32} & \textbf{137.34} \\
$\Delta$ (pp/\%)
& \cellcolor{posgreen!42}\textbf{+5.93}
& \cellcolor{posgreen!75}\textbf{-29.91\%}
& \cellcolor{posgreen!37}\textbf{+4.90}
& \cellcolor{posgreen!48}\textbf{-16.53\%}
& \cellcolor{posgreen!80}\textbf{+15.08}
& \cellcolor{posgreen!26}\textbf{-7.59\%}
& \cellcolor{posgreen!42}\textbf{+5.96}
& \cellcolor{posgreen!26}\textbf{-7.63\%} \\
\midrule

MolmoACT2
& 95.50 & 291.34
& 28.56 & 278.53
& 78.96 & 303.18
& 20.18 & 387.06 \\
\textbf{+VLA-Rolling}
& \textbf{95.70} & \textbf{196.61}
& \textbf{30.56} & \textbf{236.45}
& \textbf{79.07} & \textbf{207.27}
& \textbf{22.34} & \textbf{273.95} \\
$\Delta$ (pp/\%)
& \cellcolor{posgreen!11}\textbf{+0.20}
& \cellcolor{posgreen!80}\textbf{-32.52\%}
& \cellcolor{posgreen!21}\textbf{+2.00}
& \cellcolor{posgreen!45}\textbf{-15.11\%}
& \cellcolor{posgreen!10}\textbf{+0.11}
& \cellcolor{posgreen!78}\textbf{-31.63\%}
& \cellcolor{posgreen!22}\textbf{+2.16}
& \cellcolor{posgreen!74}\textbf{-29.22\%} \\

\bottomrule
\end{tabular}
\end{table*}

\paragraph{Evaluation Setup.} The main experiment was conducted on three simulation benchmarks: LIBERO, MetaWorld, and LIBERO-Plus. LIBERO contains 40 tasks devided into 4 suites: spatial, object, goal and long. For each task, we let the model perform 100 rollouts. MetaWorld contains 50 tasks across 4 difficulty levels, every task is evaluated with 100 rollouts. LIBERO-Plus is a perturbed version of the LIBERO dataset, incorporating common perturbations such as lighting and background variations, resulting in a total of 10,030 tasks. We test our model on every task for 1 episode. RoboTwin 2.0 contains 50 bimanual manipulation, owning corresponding variants with random perturbations to test the model's robustness. We ran 100 rollouts for each task.

\paragraph{Evaluation Results.} \Cref{tab:VLA-Rolling_results} shows that VLA-Rolling provides significant Pareto improvements, maintaining or improving task success rates while reducing inference costs. From the results, we can make two observations: First, the model's inference cost is significantly reduced. The time to infer each action chunk on different models and benchmarks is reduced by 5.00\%-32.52\%, equivalent to an increase of up to 1.48 times in action chunk throughput. This improvement comes from VLA-Rolling's adaptive denoising step estimation, which provides the VLA model with guidance for fewer denoising steps. Second, the task success rate on each benchmark remains basically unchanged, or even slightly improved. This improvement comes from lower action error at each step of action generation, resulting in less cumulative error and an increase in the number of tasks completed. These results demonstrate that VLA-Rolling provides significant Pareto improvements, maintaining or improving task success rates while reducing inference costs.

\subsection{Ablation Study}

\Cref{tab:ablation}  shows the component ablation. We can observe that simply inheriting the action tail reduced the success rate by 4.89\%, while merely adjusting the number of denoising steps also reduced it by 4.29\%. All the success rate improvements occurred in variants with renoising.

From this, we can draw two conclusions: First, even if the inherited action tail is closer to the target distribution than Gaussian noise, we must consider the characteristic that high-step denoising of the action tail impairs action quality (as mentioned in \Cref{sec:motivating_observations}). Therefore, directly inheriting the action tail and performing full-step denoising will only reduce the success rate. Second, the renoising process can significantly improve model performance. This is because the action itself sometimes lies outside the training distribution, making it difficult to correctly refine it using only the VLA's capabilities; adding a small amount of noise can reduce this possibility.

\begin{table}[t]
\centering
\caption{
Ablation study of VLA-Rolling inference components with \(\pi_{0.5}\) on LIBERO, MetaWorld, and LIBERO-Plus. A, B, C denote different components of our model. \textbf{A}: Action Tail Inheritance; \textbf{B}: Confidence-Guided Renoising; \textbf{C}: Adaptive-Step Denoising. \textbf{\(B_{\mathrm{fixed}}\)}: Fixed 50\% of renoising. The full VLA-Rolling model corresponds to A+B+C.
}
\label{tab:ablation}

\small
\setlength{\tabcolsep}{3.0pt}

\begin{tabular}{@{}lcccccc@{}}
\toprule
\multirow{2}{*}{Method}
& \multicolumn{2}{c}{\textbf{LIBERO}}
& \multicolumn{2}{c}{\textbf{MetaWorld}}
& \multicolumn{2}{c}{\textbf{LIBERO-Plus}} \\
\cmidrule(lr){2-3}
\cmidrule(lr){4-5}
\cmidrule(lr){6-7}
& SR (\%)$\uparrow$ & Cost (ms)$\downarrow$
& SR (\%)$\uparrow$ & Cost (ms)$\downarrow$
& SR (\%)$\uparrow$ & Cost (ms)$\downarrow$ \\
\midrule

$\pi_{0.5}$
& 96.38 & 166.04
& 67.22 & 217.28
& 84.73 & 161.92 \\

\midrule

\rowcolor{gray!15}
\textit{Variants}
& $\Delta$SR (pp)$\uparrow$
& $\Delta$Cost (\%)$\downarrow$
& $\Delta$SR (pp)$\uparrow$
& $\Delta$Cost (\%)$\downarrow$
& $\Delta$SR (pp)$\uparrow$
& $\Delta$Cost (\%)$\downarrow$ \\

$+A$
& \cellcolor{negyellowgreen!35}-2.13
& \cellcolor{negyellowgreen!10}+0.02\%
& \cellcolor{posgreen!14}+0.63
& \cellcolor{negyellowgreen!10}+0.01\%
& \cellcolor{negyellowgreen!50}-4.89
& \cellcolor{negyellowgreen!10}+0.02\% \\

$+A+B_{\mathrm{fixed}}$
& \cellcolor{posgreen!11}+0.20
& \cellcolor{negyellowgreen!10}+0.02\%
& \cellcolor{posgreen!14}+0.61
& \cellcolor{negyellowgreen!10}+0.02\%
& \cellcolor{negyellowgreen!15}-0.16
& \cellcolor{negyellowgreen!10}+0.02\% \\

$+A+B$
& \cellcolor{posgreen!12}+0.37
& \cellcolor{negyellowgreen!15}+0.55\%
& \cellcolor{posgreen!12}+0.40
& \cellcolor{negyellowgreen!14}+0.42\%
& \cellcolor{negyellowgreen!15}-0.22
& \cellcolor{negyellowgreen!18}+0.84\% \\

$+A+C$
& \cellcolor{negyellowgreen!30}-1.48
& \cellcolor{posgreen!60}\textbf{-22.19\%}
& \cellcolor{negyellowgreen!20}-0.62
& \cellcolor{posgreen!46}\textbf{-15.93\%}
& \cellcolor{negyellowgreen!45}-4.29
& \cellcolor{posgreen!36}\textbf{-11.91\%} \\

\midrule

\textbf{Full Model}
& \cellcolor{posgreen!18}\textbf{+1.45}
& \cellcolor{posgreen!62}\textbf{-23.03\%}
& \cellcolor{posgreen!15}\textbf{+0.66}
& \cellcolor{posgreen!48}\textbf{-16.74\%}
& \cellcolor{posgreen!14}\textbf{+0.69}
& \cellcolor{posgreen!36}\textbf{-12.07\%} \\

\bottomrule
\end{tabular}
\end{table}

\subsection{Analysis Experiments}

\paragraph{Environment Shift Study.} We explored an action confidence estimator model trained using default settings, based on $\pi_{0.5}$, and inputting environmental states at different execution time domains $K$ to simulate time shifts. The larger $K$ is, the greater the environmental change and the further it deviates from the environmental changes seen during model training. Our results in the LIBERO table \Cref{tab:observation_shift_confidence} show a negative correlation between the drasticness of environmental changes and action confidence, consistent with our assumptions about action confidence in \Cref{sec:action_confidence}.

As K increases, the correlation between confidence and environmental changes does not decrease significantly until environmental changes far exceed what the default settings (K=10) can comprehend. This indicates that action confidence estimator possesses some robustness, but still has limitations.

\begin{table}[t]
  \centering
  \caption{Observation shift and confidence statistics under different execution horizon $K$.
  Correlation values $\rho$ are Pearson correlations computed within each $K$ group. $c_{min}$ is the average minimum of predicted confidence.}
  \label{tab:observation_shift_confidence}
  \begin{tabular}{cccccc}
  \toprule
  Execution Horizon $K$
  & $\Delta o_t$
  & $\Delta q_t$
  & $c_{min}$
  & $\rho(\Delta o_t, c_{min})$
  & $\rho(\Delta q_t, c_{min})$ \\
  \midrule
  $K=$10(default) & 10.544 & 0.895 & 0.359 & $-0.257$ & $-0.237$ \\
  $K=$20 & 16.071 & 1.610 & 0.175 & $-0.318$ & $-0.207$ \\
  $K=$30 & 19.249 & 2.151 & 0.146 & $-0.272$ & $-0.200$ \\
  $K=$40 & 20.658 & 2.551 & 0.140 & $-0.100$ & $-0.155$ \\
  \bottomrule
\end{tabular}
\end{table}

\begin{figure}[!htbp]
    \centering

    % ===== Left: Figure =====
    \begin{minipage}[c]{0.55\linewidth}
        \centering
        \includegraphics[
            width=\linewidth
        ]{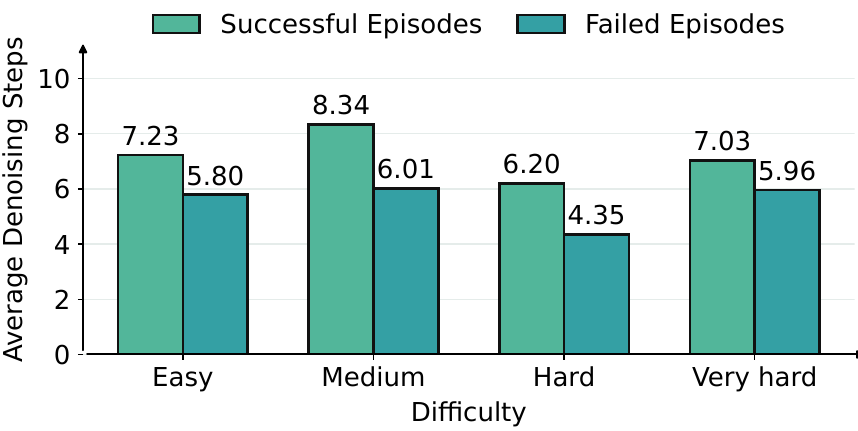}
    \end{minipage}
    \hfill
    % ===== Right: Table =====
    \begin{minipage}[c]{0.40\linewidth}
        \centering

        \begin{tabular}{lccccc}
            \toprule
            Difficulty 
            & \shortstack{Successful/ Failure}\\
            \midrule

            Easy
            & 350/70\\

            Medium
            & 87/78\\

            Hard
            & 48/42\\

            Very hard
            & 24/51\\

            \bottomrule
        \end{tabular}
    \end{minipage}

    \caption{
        \textbf{Left: Average denoising steps across task difficulty levels on MetaWorld.}
        Successful trajectories consistently require more denoising steps than failed trajectories. \textbf{Right: Distribution of episodes across different difficulty levels}
    }
    \label{fig:conf_discrimination}
\end{figure}

\paragraph{Confidence Discrimination Test.} We compared the average number of denoising steps for successful and failed tasks on MetaWorld. \Cref{fig:conf_discrimination} shows the results, revealing a significant difference in the number of denoising steps: across all task difficulty levels, successful tasks consistently required more denoising steps than failed tasks, up to 2.33 steps. This result indicates that confidence predictions significantly differentiate between successful and failed trajectories, and that denoising budget allocation is consistently related to action generation difficulty.

\begin{figure}[!htbp]
  \centering
  \includegraphics[
    width=\linewidth
  ]{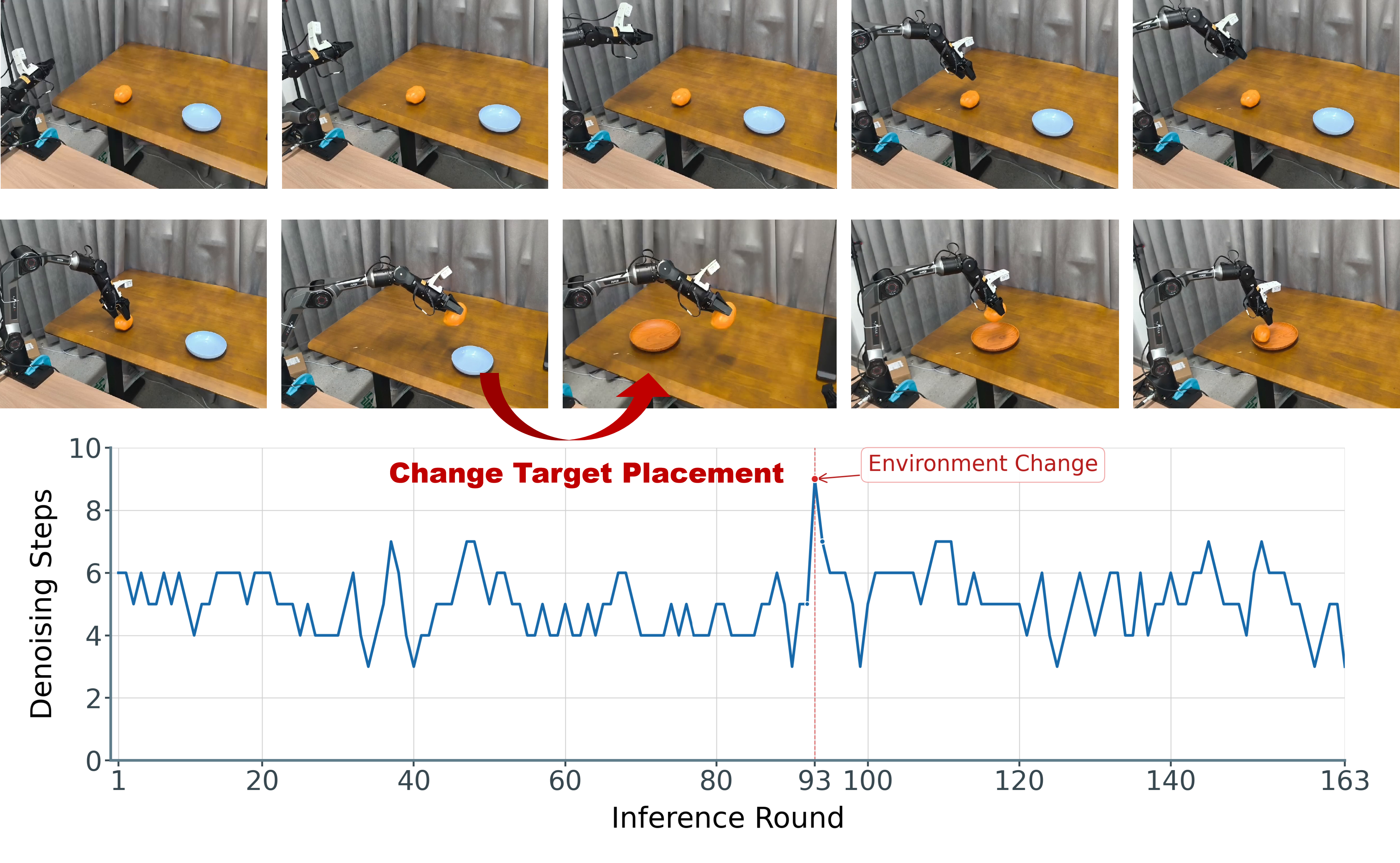}
  \caption{
    \textbf{Real-world experiment of context mutations} revealed a significant decrease in confidence at the mutation site, leading to an increase in the number of denoising steps.
  }
  \label{fig:real_world}
\end{figure}

\subsection{Real-World Experiments}

\begin{wraptable}{r}{0.43\textwidth}
\vspace{-10pt}

\centering
\small
\setlength{\tabcolsep}{3.5pt}
\caption{Model performance on real-world experiments.}
\begin{tabular}{lcc}
\toprule
Model & SR (\%) $\uparrow$ & Cost (ms)$\downarrow$ \\
\midrule
        $\pi_{0.5} $ & 70\%(14/20) & 227.28\\
        +VLA-Rolling & 75\%(15/20) & 176.50 \\
\bottomrule
\end{tabular}

\label{tab:real_world_sr}
\vspace{-10pt}
\end{wraptable}

In the real world, we used $\pi_{0.5}$ on the AgileX PiPER 6-DoF robotic arm to complete our designed target mutation task. In this task, the robotic arm needs to place potatoes into a plate. During the task, we will change the color of the target after the robotic arm approaches the target's placement position. The robotic arm then needs to continue the task to test the model's response to sudden environmental changes after VLA-Rolling is deployed. Detailed settings are shown in the Appendix.~\ref{sec:real_world_training}.

The results in \Cref{fig:real_world} show that the model initially performs the task of \textit{"putting the potato into the blue plate"} normally. However, when encountering a sudden environmental change, the confidence of the action confidence estimator prediction jumps, indicating that the confidence itself is sensitive to the environmental change. Since a decrease in confidence temporarily reverts the model to the original inference paradigm, and because the changed environment is still the same task seen during training, the model can automatically switch to the action of placing the item into the brown plate, demonstrating the practical significance of the renoising process of VLA-Rolling in the task execution process.

\section{Conclusion}

This paper investigates the reuse value of action tails that are wasted between adjacent inference rounds in the traditional VLA paradigm. We introduce VLA-Rolling, which uses a lightweight plugin to estimate the confidence of action tail, finding the optimal number of denoising steps to guide the reuse process of action tails. Our results show that VLA-Rolling can effectively predict the number of denoising steps required for action tails, efficiently generate high-quality actions, and reuse the motion information in action tails in a rolling manner in each round of inference.

\bibliography{paper}

@article{black2024pi_0,
  title={$\pi_0$: A Vision-Language-Action Flow Model for General Robot Control},
  author={Black, Kevin and Brown, Noah and Driess, Danny and Esmail, Adnan and Equi, Michael and Finn, Chelsea and Fusai, Niccolo and Groom, Lachy and Hausman, Karol and Ichter, Brian and others},
  journal={arXiv preprint arXiv:2410.24164},
  year={2024}
}

@article{intelligence2025pi_,
  title={$\pi_{0.5}$: A Vision-Language-Action Model with Open-World Generalization},
  author={Intelligence, Physical and Black, Kevin and Brown, Noah and Darpinian, James and Dhabalia, Karan and Driess, Danny and Esmail, Adnan and Equi, Michael and Finn, Chelsea and Fusai, Niccolo and others},
  journal={arXiv preprint arXiv:2504.16054},
  year={2025}
}

@article{bjorck2025gr00t,
  title={Gr00t n1: An open foundation model for generalist humanoid robots},
  author={Bjorck, Johan and Casta{\~n}eda, Fernando and Cherniadev, Nikita and Da, Xingye and Ding, Runyu and Fan, Linxi and Fang, Yu and Fox, Dieter and Hu, Fengyuan and Huang, Spencer and others},
  journal={arXiv preprint arXiv:2503.14734},
  year={2025}
}

@article{cai2026xiaomi,
  title={Xiaomi-robotics-0: An open-sourced vision-language-action model with real-time execution},
  author={Cai, Rui and Guo, Jun and He, Xinze and Jin, Piaopiao and Li, Jie and Lin, Bingxuan and Liu, Futeng and Liu, Wei and Ma, Fei and Ma, Kun and others},
  journal={arXiv preprint arXiv:2602.12684},
  year={2026}
}

@article{wang2026qwen,
  title={Qwen-vla: Unifying vision-language-action modeling across tasks, environments, and robot embodiments},
  author={Wang, Qiuyue and Li, Mingsheng and Guan, Jian and Ye, Jinhui and Xie, Sicheng and Liu, Yitao and Chen, Junhao and Liang, Zhixuan and Zhang, Jie and Hu, Xintong and others},
  journal={arXiv preprint arXiv:2605.30280},
  year={2026}
}

@article{wu2026pragmatic,
  title={A pragmatic vla foundation model},
  author={Wu, Wei and Lu, Fan and Wang, Yunnan and Yang, Shuai and Liu, Shi and Wang, Fangjing and Zhu, Qian and Sun, He and Wang, Yong and Ma, Shuailei and others},
  journal={arXiv preprint arXiv:2601.18692},
  year={2026}
}

@article{shukor2025smolvla,
  title={Smolvla: A vision-language-action model for affordable and efficient robotics},
  author={Shukor, Mustafa and Aubakirova, Dana and Capuano, Francesco and Kooijmans, Pepijn and Palma, Steven and Zouitine, Adil and Aractingi, Michel and Pascal, Caroline and Russi, Martino and Marafioti, Andres and others},
  journal={arXiv preprint arXiv:2506.01844},
  year={2025}
}

@inproceedings{zheng2026x,
  title={X-vla: Soft-prompted transformer as scalable cross-embodiment vision-language-action model},
  author={Zheng, Jinliang and Li, Jianxiong and Wang, Zhihao and Liu, Dongxiu and Kang, Xirui and Feng, Yuchun and Zheng, Yinan and Zou, Jiayin and Chen, Yilun and Zeng, Jia and others},
  booktitle={International Conference on Learning Representations},
  volume={2026},
  pages={60580--60606},
  year={2026}
}

@article{brohan2023rt,
  title={Rt-2: Vision-language-action models transfer web knowledge to robotic control},
  author={Brohan, Anthony and Brown, Noah and Carbajal, Justice and Chebotar, Yevgen and Chen, Xi and Choromanski, Krzysztof and Ding, Tianli and Driess, Danny and Dubey, Avinava and Finn, Chelsea and others},
  journal={arXiv preprint arXiv:2307.15818},
  year={2023}
}

@article{kim2024openvla,
  title={Openvla: An open-source vision-language-action model},
  author={Kim, Moo Jin and Pertsch, Karl and Karamcheti, Siddharth and Xiao, Ted and Balakrishna, Ashwin and Nair, Suraj and Rafailov, Rafael and Foster, Ethan and Lam, Grace and Sanketi, Pannag and others},
  journal={arXiv preprint arXiv:2406.09246},
  year={2024}
}

@article{lee2025molmoact,
  title={Molmoact: Action reasoning models that can reason in space},
  author={Lee, Jason and Duan, Jiafei and Fang, Haoquan and Deng, Yuquan and Liu, Shuo and Li, Boyang and Fang, Bohan and Zhang, Jieyu and Wang, Yi Ru and Lee, Sangho and others},
  journal={arXiv preprint arXiv:2508.07917},
  year={2025}
}

@article{jing2025mixture,
  title={Mixture of horizons in action chunking},
  author={Jing, Dong and Wang, Gang and Liu, Jiaqi and Tang, Weiliang and Sun, Zelong and Yao, Yunchao and Wei, Zhenyu and Liu, Yunhui and Lu, Zhiwu and Ding, Mingyu},
  journal={arXiv preprint arXiv:2511.19433},
  year={2025}
}

@article{zhao2023learning,
  title={Learning fine-grained bimanual manipulation with low-cost hardware},
  author={Zhao, Tony Z and Kumar, Vikash and Levine, Sergey and Finn, Chelsea},
  journal={arXiv preprint arXiv:2304.13705},
  year={2023}
}

@article{team2024octo,
  title={Octo: An open-source generalist robot policy},
  author={Team, Octo Model and Ghosh, Dibya and Walke, Homer and Pertsch, Karl and Black, Kevin and Mees, Oier and Dasari, Sudeep and Hejna, Joey and Kreiman, Tobias and Xu, Charles and others},
  journal={arXiv preprint arXiv:2405.12213},
  year={2024}
}

@inproceedings{o2024open,
  title={Open x-embodiment: Robotic learning datasets and rt-x models: Open x-embodiment collaboration 0},
  author={O’Neill, Abby and Rehman, Abdul and Maddukuri, Abhiram and Gupta, Abhishek and Padalkar, Abhishek and Lee, Abraham and Pooley, Acorn and Gupta, Agrim and Mandlekar, Ajay and Jain, Ajinkya and others},
  booktitle={2024 IEEE International Conference on Robotics and Automation (ICRA)},
  pages={6892--6903},
  year={2024},
  organization={IEEE}
}

@article{pertsch2025fast,
  title={Fast: Efficient action tokenization for vision-language-action models},
  author={Pertsch, Karl and Stachowicz, Kyle and Ichter, Brian and Driess, Danny and Nair, Suraj and Vuong, Quan and Mees, Oier and Finn, Chelsea and Levine, Sergey},
  journal={arXiv preprint arXiv:2501.09747},
  year={2025}
}

@article{liang2026adaptive,
  title={Adaptive action chunking at inference-time for vision-language-action models},
  author={Liang, Yuanchang and Wang, Xiaobo and Wang, Kai and Wang, Shuo and Peng, Xiaojiang and Chen, Haoyu and Chua, David Kim Huat and Vadakkepat, Prahlad},
  journal={arXiv preprint arXiv:2604.04161},
  year={2026}
}

@article{xu2026knowing,
  title={Knowing When to Stop: Adaptive Action Chunking via Internal Cross-Attention Dynamics in VLAs},
  author={Xu, Runze and Shan, Xiaolong and Dai, Shuang and Wang, Yu and Yu, Jincheng},
  journal={arXiv preprint arXiv:2609.00908},
  year={2026}
}

@inproceedings{wang2026vla,
  title={VLA Knows Its Limits: Adaptive Execution Horizons for Robot Policies},
  author={Wang, Haoxuan and Zhang, Gengyu and Yan, Yan and Kompella, Ramana Rao and Liu, Gaowen},
  booktitle={European Conference on Computer Vision},
  pages={607--623},
  year={2026},
  organization={Springer}
}

@article{black2026real,
  title={Real-time execution of action chunking flow policies},
  author={Black, Kevin and Galliker, Manuel and Levine, Sergey},
  journal={Advances in Neural Information Processing Systems},
  volume={38},
  pages={33383--33407},
  year={2026}
}

@article{black2025training,
  title={Training-time action conditioning for efficient real-time chunking},
  author={Black, Kevin and Ren, Allen Z and Equi, Michael and Levine, Sergey},
  journal={arXiv preprint arXiv:2512.05964},
  year={2025}
}

@article{tang2025vlash,
  title={Vlash: Real-time vlas via future-state-aware asynchronous inference},
  author={Tang, Jiaming and Sun, Yufei and Zhao, Yilong and Yang, Shang and Lin, Yujun and Zhang, Zhuoyang and Hou, James and Lu, Yao and Liu, Zhijian and Han, Song},
  journal={arXiv preprint arXiv:2512.01031},
  year={2025}
}

@article{lu2026faster,
  title={Faster: Rethinking real-time flow vlas},
  author={Lu, Yuxiang and Liu, Zhe and Fan, Xianzhe and Yang, Zhenya and Hou, Jinghua and Li, Junyi and Ding, Kaixin and Zhao, Hengshuang},
  journal={arXiv preprint arXiv:2603.19199},
  year={2026}
}

@article{liu2023libero,
  title={Libero: Benchmarking knowledge transfer for lifelong robot learning},
  author={Liu, Bo and Zhu, Yifeng and Gao, Chongkai and Feng, Yihao and Liu, Qiang and Zhu, Yuke and Stone, Peter},
  journal={Advances in Neural Information Processing Systems},
  volume={36},
  pages={44776--44791},
  year={2023}
}

@article{chen2025robotwin,
  title={Robotwin 2.0: A scalable data generator and benchmark with strong domain randomization for robust bimanual robotic manipulation},
  author={Chen, Tianxing and Chen, Zanxin and Chen, Baijun and Cai, Zijian and Liu, Yibin and Li, Zixuan and Liang, Qiwei and Lin, Xianliang and Ge, Yiheng and Gu, Zhenyu and others},
  journal={arXiv preprint arXiv:2506.18088},
  year={2025}
}

@inproceedings{yu2020meta,
  title={Meta-world: A benchmark and evaluation for multi-task and meta reinforcement learning},
  author={Yu, Tianhe and Quillen, Deirdre and He, Zhanpeng and Julian, Ryan and Hausman, Karol and Finn, Chelsea and Levine, Sergey},
  booktitle={Conference on robot learning},
  pages={1094--1100},
  year={2020},
  organization={PMLR}
}

@article{fei2025libero,
  title={Libero-plus: In-depth robustness analysis of vision-language-action models},
  author={Fei, Senyu and Wang, Siyin and Shi, Junhao and Dai, Zihao and Cai, Jikun and Qian, Pengfang and Ji, Li and He, Xinzhe and Zhang, Shiduo and Fei, Zhaoye and others},
  journal={arXiv preprint arXiv:2510.13626},
  year={2025}
}

@article{huang2026chainvla,
  title={ChainVLA: Chaining Vision-Language-Action Queries through a Unified Execution State for Long-Horizon Manipulation},
  author={Huang, Yuzhi and Bu, Weijue and Xiong, Ziyi and Wu, Jie and Huang, Fanding and Jiang, Jingyan and Wang, Zhi},
  journal={arXiv preprint arXiv:2608.02326},
  year={2026}
}

@article{yang2026efficientvla,
  title={Efficientvla: Training-free acceleration and compression for vision-language-action models},
  author={Yang, Yantai and Wang, Yuhao and Wen, Zichen and Zhongwei, Luo and Zou, Chang and Zhang, Zhipeng and Wen, Chuan and Zhang, Linfeng},
  journal={Advances in Neural Information Processing Systems},
  volume={38},
  pages={40891--40914},
  year={2026}
}

@article{chen2025nanovla,
  title={Nanovla: Routing decoupled vision-language understanding for nano-sized generalist robotic policies},
  author={Chen, Jiahong and Wang, Jing and Chen, Long and Cai, Chuwei and Lu, Jinghui},
  journal={arXiv preprint arXiv:2510.25122},
  year={2025}
}

@article{wen2025tinyvla,
  title={TinyVLA: toward fast, data-efficient vision-language-action models for robotic manipulation},
  author={Wen, Junjie and Zhu, Yichen and Li, Jinming and Zhu, Minjie and Tang, Zhibin and Wu, Kun and Xu, Zhiyuan and Liu, Ning and Cheng, Ran and Shen, Chaomin and others},
  journal={IEEE Robotics and Automation Letters},
  volume={10},
  number={4},
  pages={3988--3995},
  year={2025},
  publisher={IEEE}
}

@article{wang2025specprune,
  title={Specprune-vla: Accelerating vision-language-action models via action-aware self-speculative pruning},
  author={Wang, Hanzhen and Xu, Jiaming and Xiang, Yushun and Pan, Jiayi and Zhou, Yongkang and Li, Yong-Lu and Dai, Guohao},
  journal={arXiv preprint arXiv:2509.05614},
  year={2025}
}

@inproceedings{pei2026action,
  title={Action-aware dynamic pruning for efficient vision-language-action manipulation},
  author={Pei, Xiaohuan and Chen, Yuxing and Xu, Siyu and Wang, Yunke and Shi, Yuheng and Xu, Chang},
  booktitle={International Conference on Learning Representations},
  volume={2026},
  pages={10832--10851},
  year={2026}
}

@inproceedings{zhao2025cot,
  title={Cot-vla: Visual chain-of-thought reasoning for vision-language-action models},
  author={Zhao, Qingqing and Lu, Yao and Kim, Moo Jin and Fu, Zipeng and Zhang, Zhuoyang and Wu, Yecheng and Li, Zhaoshuo and Ma, Qianli and Han, Song and Finn, Chelsea and others},
  booktitle={2025 IEEE/CVF Conference on Computer Vision and Pattern Recognition (CVPR)},
  pages={1702--1713},
  year={2025},
  organization={IEEE}
}

@article{liu2025shifting,
  title={Shifting ai efficiency from model-centric to data-centric compression},
  author={Liu, Xuyang and Wen, Zichen and Wang, Shaobo and Chen, Junjie and Tao, Zhishan and Wang, Yubo and Chen, Tailai and Jin, Xiangqi and Zou, Chang and Wang, Yiyu and others},
  journal={arXiv preprint arXiv:2505.19147},
  year={2025}
}

@article{zhong2026vla,
  title={Vla-opd: Bridging offline sft and online rl for vision-language-action models via on-policy distillation},
  author={Zhong, Zhide and Yan, Haodong and Li, Junfeng and He, Junjie and Zhang, Tianran and Li, Haoang},
  journal={arXiv preprint arXiv:2603.26666},
  year={2026}
}

@article{luan2026snapflow,
  title={Snapflow: One-step action generation for flow-matching vlas via progressive self-distillation},
  author={Luan, Wuyang and Li, Junhui and Zhao, Weiguang and Zhang, Wenjian and Wu, Tieru and Ma, Rui},
  journal={arXiv preprint arXiv:2604.05656},
  year={2026}
}

@article{liu2026rdt2,
  title={Rdt2: Exploring the scaling limit of umi data towards zero-shot cross-embodiment generalization},
  author={Liu, Songming and Li, Bangguo and Ma, Kai and Wu, Lingxuan and Tan, Hengkai and Ouyang, Xiao and Su, Hang and Zhu, Jun},
  journal={arXiv preprint arXiv:2602.03310},
  year={2026}
}

@article{li2026step,
  title={STEP: Warm-started visuomotor policies with spatiotemporal consistency prediction},
  author={Li, Jinhao and Cong, Yuxuan and Wang, Yingqiao and Xia, Hao and Huang, Shan and Zhang, Yijia and Xu, Ningyi and Dai, Guohao},
  journal={arXiv preprint arXiv:2602.08245},
  year={2026}
}

@article{liu2026learning,
  title={Learning native continuation for action chunking flow policies},
  author={Liu, Yufeng and Yu, Hang and Zhao, Juntu and Li, Bocheng and Zhang, Di and Li, Mingzhu and Wu, Wenxuan and Hu, Yingdong and Xie, Junyuan and Guo, Junliang and others},
  journal={arXiv preprint arXiv:2602.12978},
  year={2026}
}

@article{sendai2025leave,
  title={Leave no observation behind: Real-time correction for vla action chunks},
  author={Sendai, Kohei and Alvarez, Maxime and Matsushima, Tatsuya and Matsuo, Yutaka and Iwasawa, Yusuke},
  journal={arXiv preprint arXiv:2509.23224},
  year={2025}
}

@article{jia2026action,
  title={Action-to-action flow matching},
  author={Jia, Jindou and Li, Gen and Chen, Xiangyu and An, Tuo and Hu, Yuxuan and Li, Jingliang and Guo, Xinying and Yang, Jianfei},
  journal={arXiv preprint arXiv:2602.07322},
  year={2026}
}

@article{jia2026mars,
  title={MARS Policy: Multimodality Only When It Matters},
  author={Jia, Jindou and An, Tuo and Hu, Yuxuan and Li, Gen and Li, Jingliang and Hou, Bohan and Chen, Xiangyu and Bai, Jiaqi and Lyu, Bofan and Yang, Jianfei},
  journal={arXiv preprint arXiv:2605.29766},
  year={2026}
}

@article{oi2026training,
  title={Training-Free Acceleration for Vision-Language-Action Models with Action Caching and Refinement},
  author={Oi, Ryuji and Otsuka, Hikari and Matsushima, Kosuke and Ichikawa, Yuki and Motomura, Masato and Kaneko, Tatsuya and Fujiki, Daichi},
  journal={arXiv preprint arXiv:2607.06370},
  year={2026}
}

@article{zhan2026seam,
  title={SEAM: Smooth Execution of Action-Chunked Motion for Vision-Language-Action Policies},
  author={Zhan, Dijia and Xu, Xuemiao and Li, Jinyi and Tang, Jie},
  journal={arXiv preprint arXiv:2607.04609},
  year={2026}
}
\bibliographystyle{assets/acl_natbib}

\clearpage
\beginappendix

\section{Related Work}
\label{sec:related_work}

\subsection{Continuous Action Generation} In embodied robot control, the VLA paradigm currently dominates~\citep{brohan2023rt,o2024open,kim2024openvla, lu2026faster, pertsch2025fast, black2024pi_0, zhao2025cot, intelligence2025pi_, bjorck2025gr00t, cai2026xiaomi, wang2026qwen, team2024octo}, utilizing rich visual language and robot motion information to generate actions. Existing paradigms use flow matching~\citep{black2024pi_0, bjorck2025gr00t, intelligence2025pi_,wu2026pragmatic, lee2025molmoact,shukor2025smolvla,zheng2026x} to transfer the priors of the Vision-Language Model as conditions to guide the generation of continuous future actions. However, only a portion of the generated actions are executed, while the remainder are discarded to prevent error accumulation. Avoiding this computational waste has become a problem that VLA models need to optimize.

\subsection{Action Chunk Reuse}

Previous work has demonstrated that information from the actions generated by the model can be used to maintain temporal consistency~\citep{black2026real,black2025training,li2026step,liu2026learning,sendai2025leave, huang2026chainvla, oi2026training, zhan2026seam}, thereby improving success rate. However, these approaches have obvious drawbacks: they rely too heavily on the quality of the action chunk itself. Low-quality action chunks are still reused as the basis for action generation, leading to incorrect action generation. Therefore, to effectively utilize action chunks, selective utilization is necessary, along with an action-level evaluation mechanism~\citep{jia2026action, jia2026mars} within each action chunk, in order to unlock the potential of actions that are discarded from the action chunk.

\subsection{Efficient VLA Inference} Efficiency optimizing paradigm for the inference of the Vision-language-Action Model varies. Some of these works attempt to address the issue from the perspective of model size~\citep{yang2026efficientvla, chen2025nanovla, wen2025tinyvla, shukor2025smolvla, wang2026vla}, requiring training the model from scratch. However, recently more work is focusing on optimizing the inference paradigm of action decoder, typically resulting in performance loss~\citep{wang2025specprune, pei2026action, liu2025shifting, zhong2026vla, luan2026snapflow, liu2026rdt2, tang2025vlash, lu2026faster}. These methods typically sacrifice the accuracy of generated actions for more efficient generation, and cannot avoid making trade-offs between the two. Few works can be universally applied to existing models and produce Patoré improvements.

\section{Method Details}
\label{sec:training_inference}

\subsection{Details of Model Architecture}

We implement the VLA-Rolling action confidence predictor as an action-centric Transformer with approximately 30M parameters, while keeping the backbone VLA frozen.

The predictor receives the current visual representation and its temporal difference, the current proprioceptive state and its temporal difference, the inherited action tail, and the last executed action. Visual token sequences are mean-pooled, and the current and difference representations are independently layer-normalized, concatenated, and projected to a 576-dimensional context vector.Finally, a Sigmoid layer is used as the final confidence output layer.

The confidence is then used by VLA-Rolling inference rules to determine the confidence-guided noise and action tail-preserving behavior.

\subsection{Details of Action Confidence Estimator}

For the simulator environment, we can directly set the execution length of each chunk, thus obtaining an action tail of a fixed length. For action confidence estimator, we strictly align the input and output lengths of its model with the deployed VLA model. For the models we use, except for the X-VLA, which uses $H=30, K=10$, and $L=20$, all other models ($\pi_{0.5}$, SmolVLA, MolmoAct2) use $H=50, K=10$, and $L=40$.

The actual input to action confidence estimator is of the same length as the chunk size $H$. Each action tail of length $L = H - K$ is padded with Gaussian noise of length $K$ before being fed into the model. During the forward pass, an attention masking mechanism is used to limit the model's focus to the action tail itself, preventing the added Gaussian noise from affecting the model's judgment of action quality. After outputting the result, it is truncated into a confidence block of length $L$ for downstream use.

To avoid having the target VLA model infer across the entire demonstration dataset to obtain the risk quantiles, we made an engineering compromise: during offline training, the $q_{0.1,i}$ and $q_{0.9,i}$ statistics are updated as exponential moving quantiles (EMA) of the valid per-position risks. Specifically, for $p\in\{0.1,0.9\}$, let $\widehat q_{p,i}^{(s)}$ be the $p$-th quantile estimated from the current training batch at update $s$; we update

\begin{equation}
    q_{p,i}^{(s)}
    =
    (1-\eta) \cdot q_{p,i}^{(s-1)}
    +
    \eta \cdot \widehat q_{p,i}^{(s)},
    \qquad p\in\{0.1,0.9\},
    \label{eq:ema_risk_quantiles}
\end{equation}

\subsection{Details of Adaptive-Step Denoising}

We consider the inherited action tail as a potential intermediate state of the flow-matching denoising trajectory, rather than restarting denoising from the fresh-Gaussian endpoint at flow time $\tau=0$. Suppose the default VLA denoising process uses $N_{\max}$ steps with step size $\delta=1/N_{\max}$. For an action tail whose optimal remaining refinement budget is $N^*$, we regard it as an intermediate state located at the corresponding remaining flow time. Specifically, letting $k=0,\dots,N^*-1$, we initialize
\begin{equation}
    \mathbf{A}_t^{(0)}
    =
    \mathbf{A}_t^{\mathrm{tail}},
    \qquad
    \tau_k
    =
    1-\frac{N^*-k}{N_{\max}},
    \label{eq:tail_flow_time}
\end{equation}
and perform the subsequent refinement using the same Euler update rule as the native VLA:
\begin{equation}
    \mathbf{A}_t^{(k+1)}
    =
    \mathbf{A}_t^{(k)}
    +
    \delta
    v_{\theta}
    \left(
        \mathbf{A}_t^{(k)},
        \tau_k
        \mid
        \mathbf{s}_t
    \right),
    \qquad
    \delta=\frac{1}{N_{\max}}.
    \label{eq:tail_flow_refinement}
\end{equation}
In this formulation, $\mathbf{A}_t^{\mathrm{tail}}$ is no longer treated as a state at $\tau=0$, but as a partially denoised intermediate state whose starting flow time is determined by its remaining refinement budget $N^*$. This preserves consistency with the timestep embedding and discretization used in the original VLA denoising process, allowing the action tail to be refined along the remaining portion of the learned flow trajectory.

\section{Training and Inference Details}

\subsection{Training Protocal}

As for the training EMA rate, we set $\eta=0.99$. We use AdamW with learning rate $10^{-4}$ and weight decay $10^{-4}$. In our L1 loss for confidence prediction, we adopt $w_{+}=2, w_{-}=1$ to realize the asymmetric penalty.

All models we adopted use batch sizes 64 for 4,000 steps. For horizon settings, we use the same default settings as the target VLA model. Training used one NVIDIA RTX PRO 6000 Blackwell Server Edition GPU with approximately 96 GiB of memory.

This online statistic provides a compact estimate of the policy- and horizon-specific risk range without retaining a large offline sample buffer or repeatedly recomputing quantiles over the full dataset. It is therefore an engineering mechanism for efficient deployment of the relative-risk normalization, rather than an additional learned prediction target. 

\subsection{Inference protocol}
\label{sec:infer_setup}

Our benchmark tests included LIBERO, MetaWorld, LIBERO-PLUS, and RoboTwin 2.0. LIBERO categorized tasks into four types, including long-range and short-range tasks. LIBERO-PLUS, as its perturbation extension, included 10,030 tasks such as camera viewpoint perturbations, used to verify the robustness of VLA-Rolling. Throughout the testing, we used a VLA model and VLA-Rolling plugin trained only on the original LIBERO demonstration dataset. MetaWorld provided four difficulty levels: Easy, Medium, Hard, and Very Hard. RoboTwin, as the only dual-arm multi-task benchmark among these, had the highest learning difficulty.

During testing, we primarily used two metrics: average success rate as a basic indicator representing model performance, and cost as an efficiency indicator, representing the inference time (in milliseconds) for each chunk. The cost, which only includes the model inference time, consists of three parts: (1) the time for the frozen VLM backbone to generate KV cache for downstream action generation, (2) the time for action expert to denoise, and (3) the forward propagation time of the VLA-Rolling confidence prediction. The reduction in adaptive denoising steps decreases the cost of part (2).

In our main experiment, we tested 100 episodes per task on LIBERO and MetaWorld, 1 episode per task on LIBERO-PLUS, and 100 episodes per task on 50 tasks in the clean subset for Robotwin. All simulator tests were conducted on four RTX 4090 32GB GPUs.

\section{Additional Experimental Results}

\subsection{Ablation Sensitivity}

We evaluated the sensitivity of VLA-Rolling to the confidence calibration parameter $\alpha$ on the LIBERO dataset. The parameter $\alpha$ controls the slope of the confidence calibration function, thus affecting how prediction confidence translates into action tail preservation and the adaptive denoising budget. We evaluated $\alpha\in\{0.0,1.0,1.5,2.0\}$ while keeping all other parameters constant. Following the same evaluation procedure, we tested all 40 tasks on the LIBERO dataset, 10 epochs per task, and reported the average success rate, average prediction denoising steps, and model inference latency per action block.

The results in \Cref{tab:alpha_ablation} show that calibration strength affects task performance and computational resource allocation. Without effective calibration ($\alpha$=0), VLA-Rolling uses an average of 6.861 denoising steps, resulting in the highest inference latency of 206.7 ms/block, with a success rate of only 97.25\%. Increasing the $\alpha$ value to 1.0 reduced the average denoising budget to 6.155 steps and inference latency to 187.6 ms/block, but the success rate slightly decreased to 97.00\%. This indicates that simply drastically reducing the denoising budget does not necessarily guarantee action quality. At $\alpha$=1.5, confidence calibration provided a more balanced distribution: the average denoising budget remained at a moderate 6.332 steps, while the success rate increased to 98.50\%. Further increasing the calibration strength to $\alpha$=2.0 did not significantly improve efficiency; instead, it reduced the success rate to 97.25\%. However, overall, VLA-Rolling remained relatively stable within the tested calibration range, consistently maintaining a success rate above 97\%.

Of all settings, $\alpha$=1.5 achieved the highest success rate of 98.50\% with an average of only 6.332 denoising steps, achieving the optimal balance between action quality and inference efficiency. Therefore, we used $\alpha=1.5$ as the default calibration parameter in all major experiments.

\begin{table}[t]
  \centering
  \caption{Sensitivity of VLA-Rolling performance to the calibration parameter $\alpha$ on LIBERO.
  We test 40 tasks with 10 episodes per task.}
  \label{tab:alpha_ablation}
  \begin{tabular}{cccc}
  \toprule
  $\alpha$
  & Success Rate (\%)
  & Average Denoising Steps
  & Model Inference (ms/chunk) \\
  \midrule
  0.0
  & 97.25
  & 6.861
  & 206.7 \\
  1.0
  & 97.00
  & 6.155
  & 187.6 \\
  1.5 (default)
  & \textbf{98.50}
  & 6.332
  & 195.7 \\
  2.0
  & 97.25
  & 6.349
  & 190.9 \\
  \bottomrule
\end{tabular}
\end{table}

\subsection{Ablation on Denoising Step Convert}

We tested different ways of converting confidence to denoising steps and calculated their trajectory-level cumulative errors. The results in \Cref{tab:reduce_trajectory_error} showed that, overall, taking the minimum confidence as the basis for denoising steps better meets our need to avoid sacrificing accuracy, so we selected it as the default setting.

\begin{table}[t]
  \centering
  \caption{Offline trajectory-level action error under different denoising step convert strategies.}
  \label{tab:reduce_trajectory_error}
  \begin{tabular}{llccc}
  \toprule
  Benchmark & variation & Mean Trajectory Error & Std. & Avg. Denoising Steps \\
  \midrule
  \multirow{3}{*}{LIBERO}
  & min  & 53.739 & 17.311 & 6.33 \\
  & mean & 54.420 & 17.964 & 4.36 \\
  & max  & 55.989 & 15.979 & 2.77 \\
  \midrule
  \multirow{3}{*}{MetaWorld}
  & min  & 6.046 & 2.772 & 7.23 \\
  & mean & 6.165 & 2.909 & 5.63 \\
  & max  & 6.585 & 2.953 & 4.12 \\
  \bottomrule
\end{tabular}
\end{table}

\subsection{Ablation on action confidence estimator parameters}

We investigate the impact of VLA-Rolling model capacity by varying the size of the confidence estimator, while keeping the VLA backbone and all other inference settings constant. We evaluate three variants on the LIBERO dataset with approximately 10M, 30M, and 100M parameters, respectively. Following the same evaluation protocol, all 40 tasks were tested, with each task running for 10 epochs. We report the average success rate and its variation across different tasks, as well as the inference cost introduced by VLA-Rolling itself.

As shown in \Cref{tab:VLA-Rolling_libero_cost}, increasing the estimator size from 10 million to 30 million improves the average success rate from 97.25\% to 98.00\%, while the inference cost only increases slightly, from 1.267 ms to 1.579 ms. In contrast, scaling the model to 100M parameters did not yield additional performance improvements; instead, it increased inference cost to 1.731 ms. This indicates that a moderate model size is sufficient to capture the relationship between inherited action tails and the updated execution context, and further increasing the model size provides limited benefits. It also demonstrates that the effectiveness of VLA-Rolling primarily stems from its action tail reuse mechanism, rather than simply increasing model size. Therefore, we used a 30M parameter configuration as the default VLA-Rolling model in all major experiments.

\begin{table}[t]
  \centering
  \caption{Performance and inference cost of VLA-Rolling models on LIBERO.
  Success rate statistics are computed across 40 tasks, 10 episodes per task
  }
  \label{tab:VLA-Rolling_libero_cost}
  \begin{tabular}{lcc}
  \toprule
  \textbf{VLA-Rolling Model} &
  \textbf{Success Rate (\%)} &
  \textbf{VLA-Rolling Cost (ms)} \\
  \midrule
  10M  & $97.25 \pm 5.47$ & $1.267 \pm 0.105$ \\
  30M  & $98.00 \pm 5.10$ & $1.579 \pm 0.335$ \\
  100M & $97.25 \pm 7.41$ & $1.731 \pm 0.047$ \\
  \bottomrule
\end{tabular}
\end{table}

\subsection{Ablation on N* Nearby Permutation}

\begin{table}[t]
    \caption{\textbf{N* Locally Optimal Test.} }
    \centering
    \begin{tabular}{lccc}
    \toprule
    Benchmark
    & $N-1$$\uparrow$
    & \textbf{$N$}$\uparrow$
    & $N+1$$\uparrow$ \\
    \midrule
    
    LIBERO
    & 97.00 {\scriptsize($-0.83$)}
    & \textbf{97.83}
    & 96.75 {\scriptsize($-1.08$)} \\
    
    MetaWorld
    & 66.67 {\scriptsize($-1.21$)}
    & \textbf{67.88}
    & 67.27 {\scriptsize($-0.61$)} \\
    \bottomrule
    \end{tabular}
    \label{tab:N_optimal}
\end{table}

To further examine whether the adaptive denoising budget obtained from the VLA-Rolling confidence linear mapping reflects the actual refinement requirements of the inherited action tail, we perturbed the prediction step size by $\pm1$. As shown in \Cref{tab:N_optimal}, both perturbations consistently degrade task performance. On LIBERO, changing the predicted budget to $N-1$ and $N+1$ reduces the success rate by $0.83$ and $1.08$ percentage points, respectively. A similar trend is observed on MetaWorld, where the corresponding drops are $1.21$ and $0.61$ percentage points. This local sensitivity suggests that the budget predicted by VLA-Rolling is closely aligned with the refinement demand of individual action tails, rather than behaving as an arbitrary reduction in denoising steps.

\subsection{Environmental similarity statistics}

\label{sec:env_sim}

\begin{table}[t]
  \centering
  \caption{\textbf{Observation and proprioceptive-state similarity across a 10-step execution horizon.}
    We compare the source context at time $t$ with the updated context at time $t+10$.
    \textbf{Image Cosine} denotes the cosine similarity between the two visual representations;
    \textbf{Image $(1-\mathrm{MAE})$} denotes one minus the mean absolute error between the normalized visual representations;
    \textbf{State Cosine} measures the cosine similarity between the corresponding proprioceptive states
    }
  \label{tab:observation_similarity}
  \begin{tabular}{lcccc}
  \toprule
  Benchmark
  & Image Cosine
  & Image $(1-\mathrm{MAE})$
  & State Cosine
  & Standardized State Distance \\
  \midrule
  LIBERO
  & $0.953 \pm 0.028$
  & $0.951 \pm 0.019$
  & $0.998 \pm 0.003$
  & $0.309 \pm 0.222$ \\
  MetaWorld
  & $0.992 \pm 0.007$
  & $0.986 \pm 0.009$
  & $0.991 \pm 0.012$
  & $0.506 \pm 0.307$ \\
  \bottomrule
\end{tabular}
\end{table}

These statistics indicate that refreshing observations after performing $K=10$ steps typically does not correspond to entirely different environmental states. While rolling in the time domain remains necessary to incorporate newly observed biases and prevent error accumulation, the high similarity between adjacent contexts suggests that previously generated action tails are not necessarily completely invalidated after replanning. This provides an empirical basis for VLA-Rolling, enabling the selective reuse of information contained in action tails, rather than unconditionally discarding them and regenerating the entire action chunk from new Gaussian noise.

\section{Real-World Details}

\subsection{Real-World Experiment Setting}

\label{sec:real_world_training}

In our real-world experiments, we chose to use the open-source \texttt{pi05-{base}} from Lerobot as the base model, the AgileX PiPER 6-DoF robotic arm as the end effector, and designed a target placement mutation task to address the confidence dependence on environmental changes.

This task consists of two phases: the robotic arm grasps a potato and attempts to place it into a blue plate. When the potato approaches the upper plate, a perturbation is introduced, removing the blue plate and replacing it with a brown plate. The robotic arm must continue its action and place the potato into the brown plate to complete the task.

We deliberately ambiguous the language instruction for this task as "place the potato into the plate" without specifying the color, allowing both modes to share the same linguistic prior. We then operated the robotic arm to collect 20 trajectories each from the blue and brown plates as target placement points. First, we fine-tuned the \texttt{pi05-{base}} model on the target data, and then used the fine-tuned pi05 model as the frozen VLA backbone to train action confidence estimator.

\subsection{Real-World Training Protocol}

During real-world deployment, the adaptation of asynchronous execution and VLA-Rolling needs to be considered: the execution horizon of asynchronous execution is determined by both inference latency and control frequency. However, under the control of adaptive denoising steps, our inference latency is not a constant value. Since the time consumed by flow matching accounts for a relatively fixed proportion in the denoising process and is basically linearly related to the number of denoising steps, the inference latency can be estimated from the inference time of its various components.

According to the definition of \Cref{tab:VLA-Rolling_results}, we can formally calculate the inference latency $C$ as:

\begin{equation}
    C_{max} = C_{FM} + C_{VLARP} + C_{non-FM}
\end{equation}
\begin{equation}
    C_{min} = \frac{1}{N_{full}}\cdot C_{FM} + C_{VLARP} + C_{non-FM}
\end{equation}

where $C_{FM}$ and $C_{VLARP}$ denotes the forward computation cost of Action Expert and action confidence estimator, $C_{non-FM}$ denotes the cost of the rest of parts.

Subsequently, we can estimate the approximate distribution of the execution horizon:

\begin{equation}
    K_{min} = \lfloor C_{min} \cdot f_{control} \rfloor
    \qquad
    K_{max} = \lceil C_{max} \cdot f_{control} \rceil
\end{equation}

where the $f_{control}$ denotes the control frequency.

For general models, the upper and lower bounds of this distribution will not exceed 100ms in physical time, making it difficult to generate excessive environmental changes. Therefore, we choose to construct data pairs for training on these execution horizons. Finally, we train for one epoch and put it into use.

\section{Limitation and Future Work}

This work has limited exploration of cross-environment generalization. Experiments on real robots can be extended to more complex task settings to explore factors beyond environmental abrupt changes that can influence confidence estimation, providing more evidence to explain the mechanism of confidence prediction. Broadening real-world evaluation across additional embodiments and interaction settings is a natural next step. Furthermore, the combination of our inference paradigm with other inference acceleration mechanisms is worth exploring, as this work does not require any modification to the original model and has the potential to accelerate model inference in conjunction with other mechanisms.

\end{document}